\documentclass[pdflatex,sn-mathphys-num]{sn-jnl}

\usepackage{graphicx}%
\usepackage{multirow}%
\usepackage{amsmath,amssymb,amsfonts}%
\usepackage{amsthm}%
\usepackage{mathrsfs}%
\usepackage[title]{appendix}%
\usepackage{xcolor}%
\usepackage{textcomp}%
\usepackage{manyfoot}%
\usepackage{booktabs}%
\usepackage{algorithm}%
\usepackage{algorithmicx}%
\usepackage{algpseudocode}%
\usepackage{listings}%

\theoremstyle{thmstyleone}%
\theoremstyle{thmstyletwo}%

\theoremstyle{thmstylethree}%

\begin{document}

\title[Article Title]{Exploring Adversarial Watermarking in Transformer-Based Models: Transferability and Robustness Against Defense Mechanism for Medical Images}


\author[1]{\fnm{Rifat} \sur{Sadik}}\email{rifatsdk@udel.edu}
\equalcont{These authors contributed equally to this work.}

\author[1]{\fnm{Tanvir} \sur{Rahman}}\email{rtanvir@udel.edu}
\equalcont{These authors contributed equally to this work.}

\author[1]{\fnm{Arpan} \sur{Bhattacharjee}}\email{arpan@udel.edu}

\author[2]{\fnm{Bikash} \sur{Chandra Halder}}\email{bkashju@gmail.com}
\author[3]{\fnm{Ismail} \sur{Hossain}}
\email{ihossai4@gmu.edu}

\author[4]{\fnm{Mridul} \sur{Banik}}\email{mridul.banik23@alumni.colostate.edu}

\author*[5]{\fnm{Jia} \sur{Uddin}}\email{jia.uddin@wsu.ac.kr}

\affil[1]{\orgdiv{Computer and Information Sciences}, \orgname{University of Delaware}, \orgaddress{\city{Newark}, \postcode{19711}, \state{Delaware}, \country{USA}}}

\affil[2]{\orgdiv{Computer Science and Engineering}, \orgname{Jahangirnagar University}, \orgaddress{\city{Dhaka}, \postcode{1342}, \country{Bangladesh}}}

\affil[3]{\orgdiv{Department of Computer Science}, \orgname{George Mason University}, \orgaddress{\ \city{Fairfax}, \postcode{22030}, \state{Virginia}, \country{USA}}}

\affil[4]{\orgdiv{Department of Computer Engineering}, \orgname{Chosun University}, \orgaddress{\ \city{Dong-gu,}, \postcode{61452}, \state{Gwangju}, \country{Republic of Korea}}}

\affil*[5]{\orgdiv{Artificial Intelligence and Big Data Department}, \orgname{Endicott College, Woosong University}, \orgaddress{\ \city{Daejeon},  \country{Republic of Korea}}}


\abstract{

Deep learning models have shown remarkable success in dermatological image analysis, offering potential for automated skin disease diagnosis. Previously, convolutional neural network (CNN)-based architectures achieved immense popularity and success in computer vision (CV) tasks such as skin image recognition, generation, and video analysis. However, with the emergence of transformer-based models, many CV tasks are now being carried out using these models. Vision Transformers (ViTs) are transformer-based models that have shown strong performance in computer vision. They use self-attention mechanisms to achieve state-of-the-art results across various tasks. However, their reliance on global attention mechanisms makes them susceptible to adversarial perturbations. This paper aims to investigate the susceptibility of ViTs in medical imaging to adversarial watermarking—a method that adds imperceptible perturbations to fool models. By generating adversarial watermarks through Projected Gradient Descent (PGD), we examine the transferability of such attacks to CNNs and analyze the effectiveness of a defense mechanism: adversarial training. Results indicate that while performance is not compromised for clean images, ViTs become much more vulnerable to adversarial attacks, with an accuracy drop as low as 27.6\%. Nevertheless, adversarial training raises it to 90.0\%.}




\maketitle

\section{Introduction}


Artificial Intelligence has helped advance dermatological practice by automatically and accurately classifying skin lesions to support clinical decision-making, thereby expanding teledermatology. We can compare the performance of convolutional neural networks (CNNs), among other deep learning models, with that of industry experts when it comes to the recognition and diagnosis of skin cancers, inflammatory dermatoses, and other dermatological conditions. Due to their ability to capture global contextual relationships across images, Vision Transformers (ViTs) have gained credibility as a promising alternative to CNNs. This is especially relevant in dermatology because accurate results often require global attention. Subtle but crucial diagnostic characteristics frequently extend beyond local anatomical boundaries and play an important role in classification.

Transfer-based architectures, originally designed for natural language processing, have recently shown promise in the field of image processing. Vision Transformers (ViTs), with their reliance on self-attention mechanisms, have demonstrated superior performance in tasks such as image recognition, segmentation, and multimodal learning \cite{mao2022towards}. As the medical domain is known for the scarcity of data, transfer-based architectures will also play an important role in this field in terms of segmentation, anomaly detection, and eventually image recognition, i.e., diagnosis. Unlike traditional Convolutional Neural Networks (CNNs) that process localized spatial features, ViTs operate on global relationships across image patches, allowing them to capture long-range dependencies effectively. This enables them to outperform CNNs in specific scenarios, including subtle pattern identification in complex medical images. However, this global attention mechanism also exposes them to unique vulnerabilities, particularly in adversarial settings.

\begin{figure}
    \centering
    \includegraphics[width=0.8\linewidth]{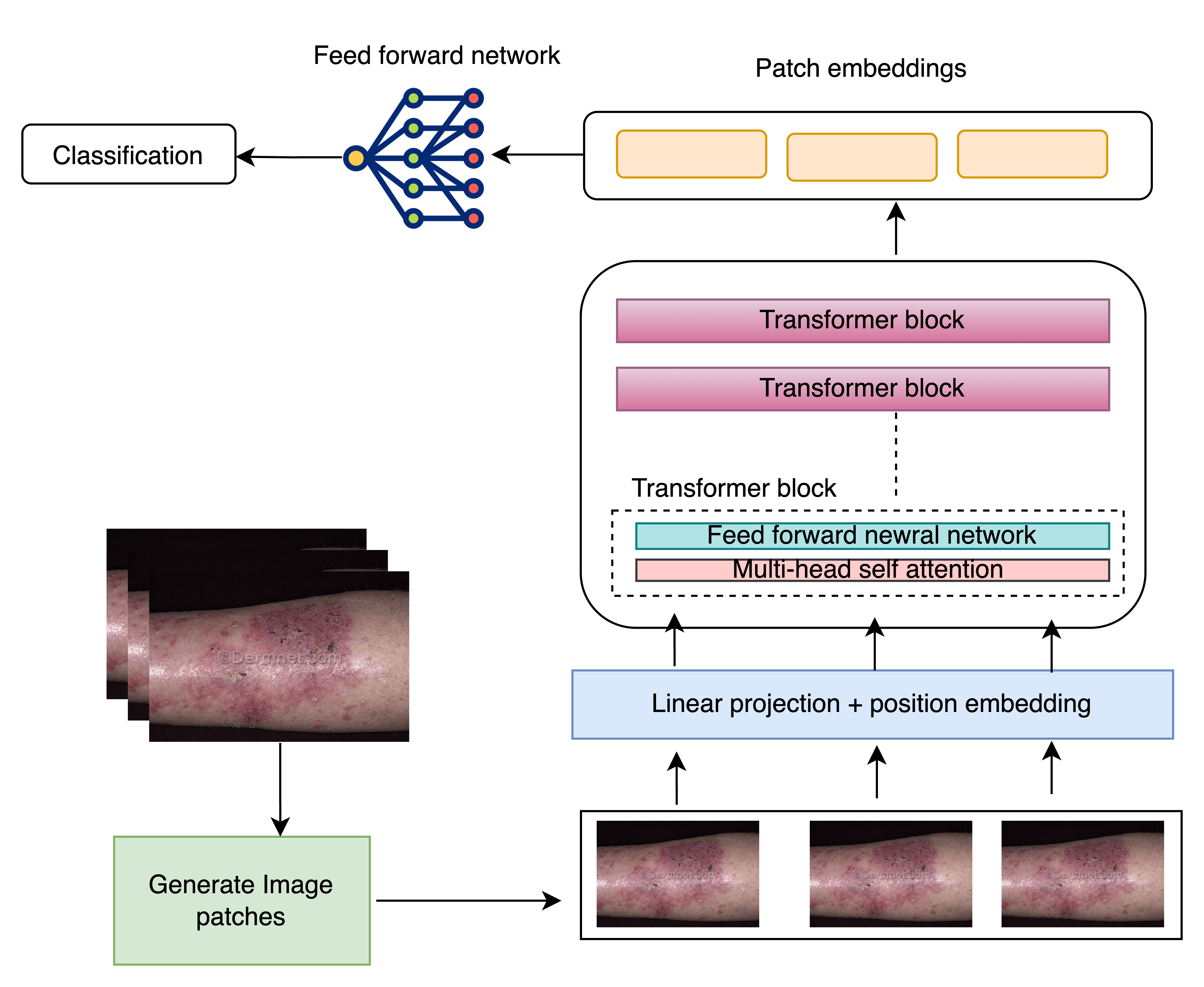} 
    \caption{Vision Transformer}
    \label{fig:vit}
\end{figure}

It’s crucial to understand the architecture of ViTs, since their vulnerabilities lie in the architecture.  Figure \ref{fig:vit} represents the Vision Transformer (ViT) Architecture\footnote{\url{https://towardsdatascience.com/using-transformers-for-computer-vision-6f764c5a078b}}, which is used for image classification tasks utilizing the transformer mechanism. First, the input image is divided into smaller fixed-size patches (e.g., 16×16). Each image patch is flattened and passed through a linear projection layer to create patch embeddings. Positional embeddings are added to the patch embeddings to retain spatial information. A sequence of transformer blocks processes the patch embeddings. Each block includes: Multi-Head Self-Attention: Captures global relationships between patches and Feed Forward Neural Network: Enhances the representation of each patch embedding. The processed embeddings from the final transformer block represent the entire image. The final embedding is passed through a feed-forward network for classification into target classes.


Adversarial vulnerability plays a vital role in clinical dermatology. If not detected by clinicians and patients, even a subtle change can lead to major diagnostic errors. It may cause misclassifications of malignant melanoma as benign melanoma or vice versa. Therefore, these misclassifications could lead to inadequate treatment planning, delayed diagnoses, and unnecessary procedures that will affect patients and their experiences. 

ViTs take images—in this case, medical images—divide them into patches of a fixed size, and then map these patches linearly into embeddings. This patch-based representation lacks the kind of hierarchical feature extraction that even the simplest CNNs perform, where lower layers recognize simple shapes and higher layers recognize much more complex features. In contrast, the attention mechanism in ViTs considers all patches simultaneously when computing their relationships. Therefore, perturbations that change the interrelationships among patches can be used to cause misrepresentation of attention maps with respect to the original structure of the image. ViTs also operate with global attention mechanisms, which means they are more vulnerable to small, subtle perturbations that are distributed globally. Transferability is an important aspect of adversarial attacks, where perturbations designed for one model can effectively impact another model, even if they differ architecturally. Research indicates that both Vision Transformers and CNNs exhibit this vulnerability, highlighting the need for robust defense strategies. For example, Liao et al. elaborated on how adversarial examples can be transferred across different model architectures \cite{liao2018defense}, while Xu et al. showed the effectiveness of various defense methods against these attacks, indicating that adversarial training can fortify models, including both CNNs and transformer-based architectures \cite{xu2021towards}.

\begin{figure}
    \centering
    \includegraphics[width=0.8\linewidth]{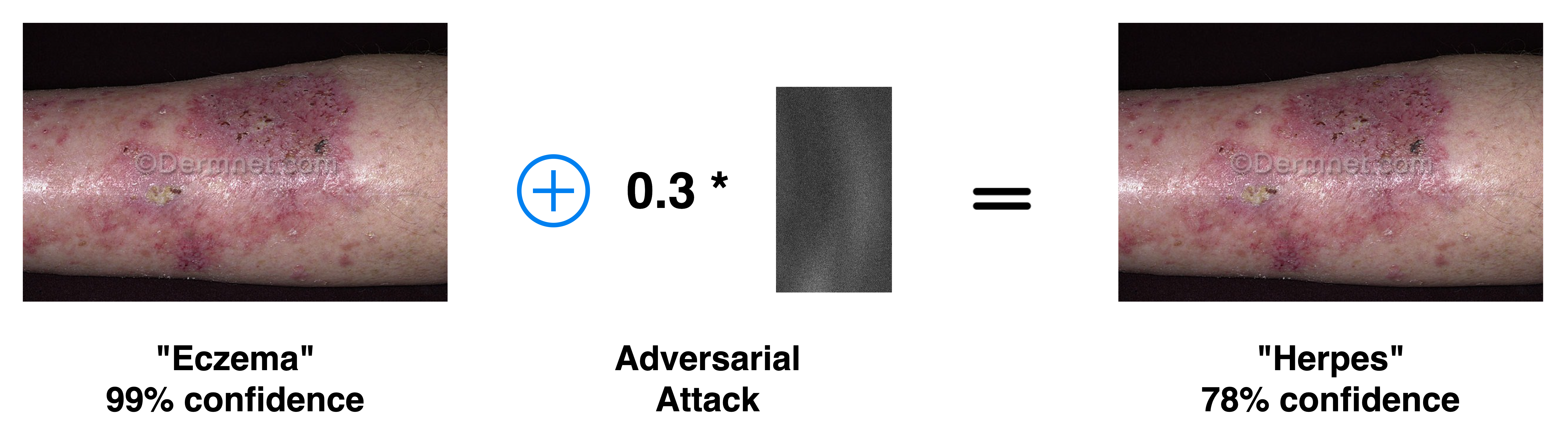}
    \caption{Adversarial watermark generation}
    \label{fig:AW}
\end{figure}

The main purpose of this research is to analyze the susceptibility of Vision Transformers (ViTs) to adversarial attacks and the transferability of these attacks to different architectures, such as Convolutional Neural Networks (CNNs). For these experiments, Projected Gradient Descent (PGD) will be used as an adversarial watermarking method to create perturbations on images in order to generate adversarial examples for ViTs. In addition, the study will examine whether such adversarial watermarks can be transferred across model architectures, as they represent a threat in heterogeneous deep learning environments. To address these gaps, the research will then explore and evaluate a variety of defense mechanisms against these vulnerabilities, such as adversarial training, to make ViTs more robust against adversarial perturbations. In summary, these goals should further the broader understanding of contemporary adversarial threats to deep learning architectures and potential avenues for robust defenses.

\section{Related Works}

\subsection{Adversarial Attacks and Watermarking in Deep Learning}

Adversarial attacks pose significant threats to the integrity of deep learning models, particularly in sensitive applications such as medical imaging. The strategy of embedding watermarks has emerged as a method for securing digital assets against such attacks. Watermarking in deep learning is used to authenticate the source and ownership of images, ensuring that any unauthorized alterations can be detected \cite{Kaczyński, Mareen}. In medical contexts, where data integrity is a must, these watermarking techniques not only protect the authenticity of the images but also deal with any potential security breaches \cite{LiDekai}. The need for robust watermarking systems grows in importance as adversarial attacks become more sophisticated, underscoring their essential role in security strategies within the healthcare domain \cite{Clements}.

\subsection{Traditional CNN-based Watermarking}

Traditional watermarking techniques often leverage Convolutional Neural Networks (CNNs) to embed invisible marks into digital media. Approaches such as invisible embedding and frequency domain methods have been widely studied, where information is embedded in transformed image representations \cite{WuShiqiang, Dong}. However, these methods frequently encounter limitations regarding robustness and detection, as they may be susceptible to various forms of image manipulations, such as geometric transformations and noise, which impair watermark visibility and durability \cite{Jabra}. Despite addressing these issues to some extent, the inherent fragility of CNN-based schemes highlights the ongoing need for improvements in robustness \cite{ZhangWenxing}.

\subsection{Transformer-based Adversarial Watermarking}

Recent advancements have begun to explore transformer architectures for watermarking within medical image processing. Research indicates that transformers provide enhanced feature extraction capabilities compared to traditional CNNs, facilitating the handling of diverse image characteristics encountered in medical datasets \cite{LiuJing}. This shift not only leads to potentially better watermarking techniques but also offers a comparative advantage in versatility and robustness against various attack vectors \cite{SinghHimanshu}. Current studies suggest that the integration of transformer models can significantly enhance watermark reliability, although further empirical investigations are required to fully understand their efficacy in real-world applications \cite{Mahapatra, PuBangzheng}.

\subsection{Transferability of Adversarial Watermarks}

Transferability of adversarial watermarks between different models and datasets has emerged as a critical research area, particularly in the context of deep learning. Existing literature underscores the complexities involved in ensuring that watermarks remain intact across various architectures and data types \cite{Zhong}. However, there is a notable gap in studies specifically focusing on transformer watermark transferability in medical imaging domains, leaving a crucial area unexplored \cite{LiuXinwei}. This lack of targeted research limits the understanding of how adversarial watermarks can function effectively in diverse model environments, particularly within healthcare applications where data migration and interoperability are prevalent challenges \cite{review}.

\subsection{Medical Imaging Specific Context}

In the medical imaging field, watermarking techniques are particularly sensitive to issues of security, ethical implications, and compliance with data protection regulations. Research has illustrated that certain watermarking methodologies designed for medical modalities (e.g., MRI, CT) face unique challenges, necessitating a tailored approach to ensure both invisibility and robustness \cite{Nie, SinghKamred}. The integration of advanced watermarking solutions, such as using patient-specific data as a watermark, has been suggested as a means to bolster both security and ethical standards in healthcare \cite{Zhang}. Nevertheless, the application of these methods demands thorough validation to align with stringent medical standards concerning data integrity and patient privacy \cite{LiuXinwei}.

Despite advancements in adversarial watermarking, significant gaps persist in the literature. There is a noticeable lack of extensive studies addressing watermark transferability specifically with transformer models. That limits their insights into their utility across varied healthcare applications. Moreover, insufficient robustness testing against advanced defensive techniques raises concerns about the practical viability of current methods in real-world scenarios \cite{SinghKamred}. Finally, limited practical experimentation using transformer architectures within prevalent medical imaging datasets underscores the need for focused research to fill these gaps and improve security in medical imaging systems.


\section{Methodology}
\subsection{Projected Gradient Descent (PGD)}

PGD creates adversarial perturbations that are small but effective at deceiving a model, causing it to misclassify the input. The goal is to maximize the model’s prediction loss while keeping the perturbation imperceptible. The update rule for PGD-based adversarial watermark generation is:

\begin{equation}
x_{t+1}' = \Pi \left( x_t + \alpha \cdot \text{sign} \left( \nabla_x J(\Theta, x_t, y) \right) \right)
\end{equation}

where \( x_t \) is the input at iteration \( t \), \( \alpha \) is the step size, \(\nabla_x J(\Theta, x_t, y)\) represents the gradient of the loss function with respect to the input, and \(\Pi\) is the projection operator that ensures the perturbed input stays within predefined bounds.

\begin{figure}
    \centering
    \includegraphics[width=0.6\linewidth, angle=90]{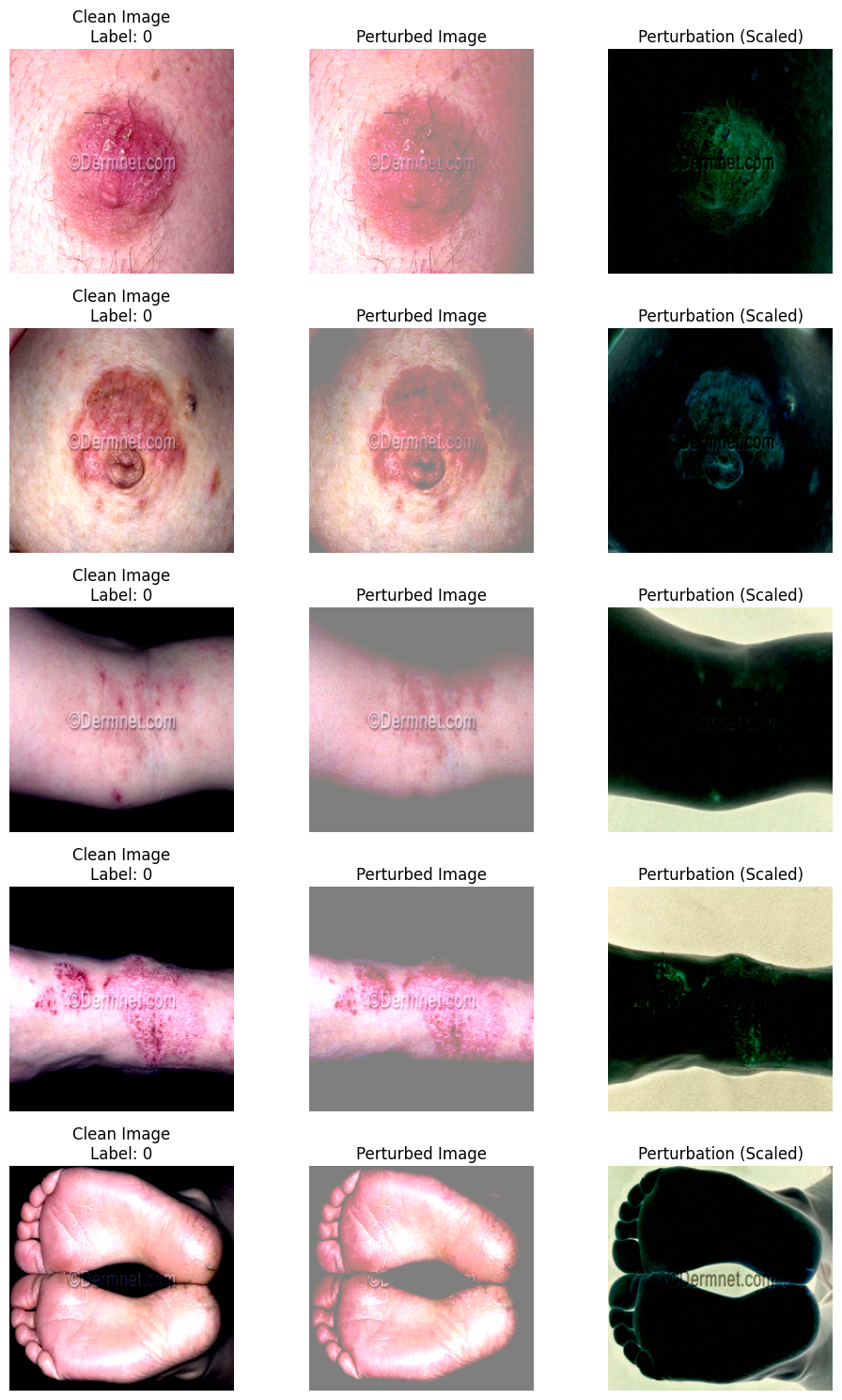}
    \caption{Images generated using PGD adversarial images}
    \label{fig:Watermarked}
\end{figure}

The PGD algorithm basically computes the loss function gradient. This gradient will divulge which direction the image should be slightly varied so that it misleads the model. It generates a small perturbation step-by-step through this gradient towards the image. Then the PGD adds a tiny perturbation to it (that's regulated by $\alpha$ in step size) in each step. Then, after the perturbation has been added, PGD ensures containment of the modified image within certain limits, as in the boundary box confined by $\epsilon$. This makes sure that the perturbations are so small that they are not visible to a human. PGD has a set number of iterations like that. This makes the image "broken" for the model with each step, although people see it as similar to the original. After several iterations, PGD yields an excellent image that seems real to the naked eye but is, according to the model being used, a different one.

In figure \ref{fig:Watermarked}, it can be clearly observed that the clean and perturbed images look nearly identical to humans, confirming that adversarial attacks are subtle and do not visibly distort the image. The scaled perturbations often highlight specific regions (e.g., edges, textures) that are critical to the model's decision-making process.

\subsection{Proposed Workflow}
\begin{figure}[htb]
    \centering
    \includegraphics[width=\linewidth]{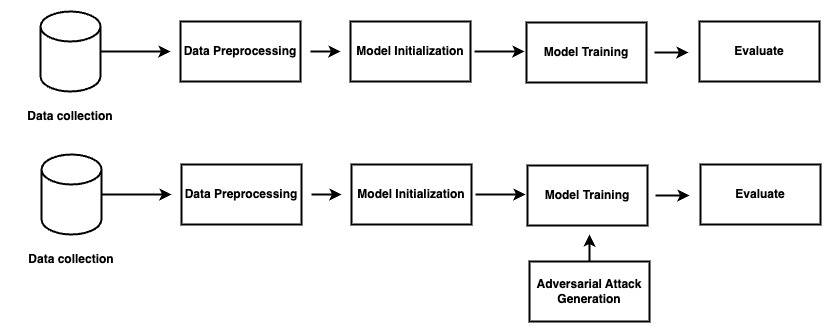}
    \caption{Workflow of our proposed approach}
    \label{fig:Workflow}
\end{figure}

In figure \ref{fig:Workflow}, the proposed workflows are given. Data collection for the standard workflow of the training ViT model, wherein the data set is collected for training, follows these preprocessing steps (normalization and augmentation), which prepares the data for the model before initiating the architecture of the model and training it on the precomputed progressive data. Finally, the evaluation takes place through a validation or test set. The training process after that is similar; however, it goes ahead to add adversarial perturbations into the dataset to create adversarial attacks at the end of training the model, thus strengthening the model's robustness. The verification exercise of the model performance will then be done both under normal conditions and during adversarial conditions to measure its performance under attack.

\section{Experiment and Results}

\subsection{Dataset and Preprocessing}

We used five classes of skin diseases: Atopic dermatitis, Eczema, Herpes, Nevus, and Melanoma. This dataset was used in a study conducted by Sadik et al. \cite{sadik2023depth}, in which the authors proposed an automated system for skin disease recognition using CNN architectures.

For data preprocessing, images were resized to  224×224 to match the input requirements of Vision Transformers (ViTs).
Additional preprocessing steps included normalization to standardize pixel values. Random cropping and flipping for data augmentation are also performed.

The training and evaluation process is performed in Google Colab, which provides GPU access.  The training process was configured with the following parameters:
\begin{itemize}
    \item Optimizer: Adam
    \item Learning rate: 0.0001
    \item Batch size: 32
\end{itemize}

\subsection{Results and Discussions}

This section presents the empirical findings of our adversarial robustness experiments, which were conducted on a skin disease image classification dataset. The effectiveness of three deep learning methods is justified with both clean and adversarial conditions. Also, the impact of adversarial training on improving the models' robustness is analyzed. 

\begin{figure}
    \centering
    \includegraphics[width=0.8\linewidth]{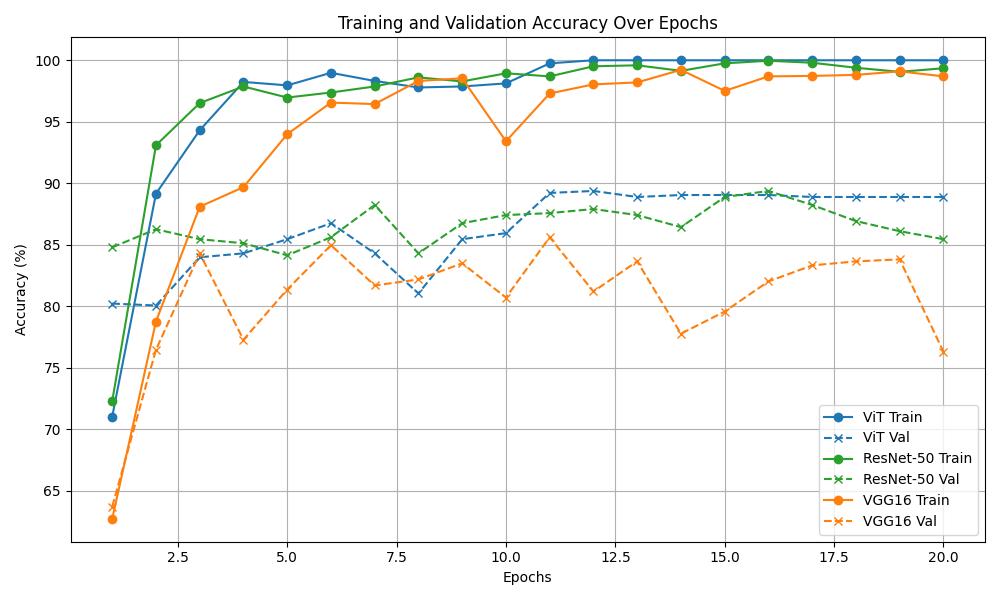}
    \caption{Training and validation accuracy of ViT, ResNet-50, and VGG16 over 20 epochs on the skin disease image dataset.}
    \label{fig:clean_aacuracy}
\end{figure}

First, three models were trained on clean images to compare baseline classification accuracy in Figure \ref{fig:clean_aacuracy}. In this experiment, ViT and ResNet-50 delivered accuracies of 94.4\% and 94.2\%, respectively, while VGG16 achieved 84.0\%. These results demonstrate the effectiveness of transformer-based and deep CNN architectures in learning discriminative features for skin disease classification.

\begin{figure}
    \centering
    \includegraphics[width=0.8\linewidth]{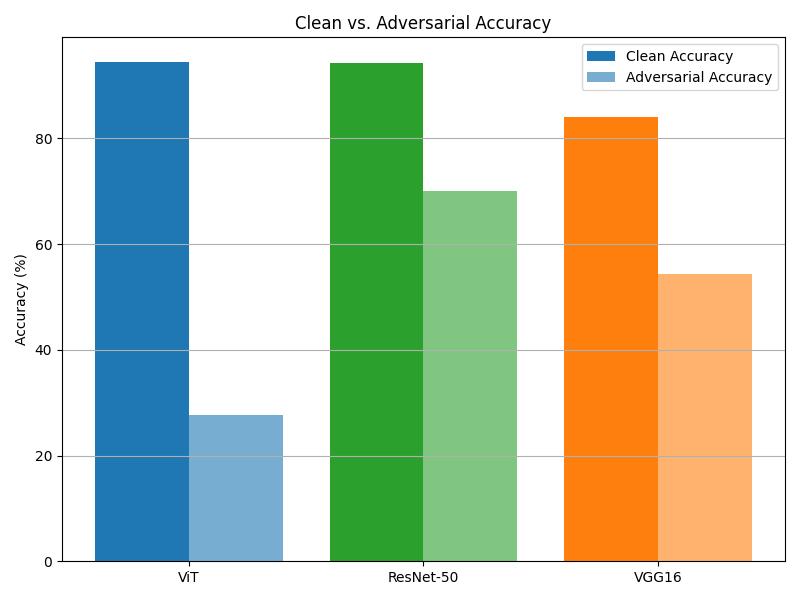}
    \caption{Comparison of classification accuracy on clean versus adversarial examples for ViT, ResNet-50, and VGG16.}
    \label{fig:Adversarial_Image_aacuracy}
\end{figure}

To verify the robustness of models, adversarial examples were crafted using the Projected Gradient Descent (PGD) method, and the results are shown in figure \ref{fig:Adversarial_Image_aacuracy}. During the attack, ViT performed very poorly with an accuracy of 27.6\%, while ResNet-50 and VGG16 degraded to 70.0\% and 54.4\%, respectively. Therefore, it can be said that ViT is much more vulnerable to adversarial attacks due to the use of global self-attention that makes it sensitive to small pixel changes. But ResNet-50 and VGG16 are more resistant to these attacks because they focus on local features, making them more robust.

\begin{figure}
    \centering
    \includegraphics[width=0.8\linewidth]{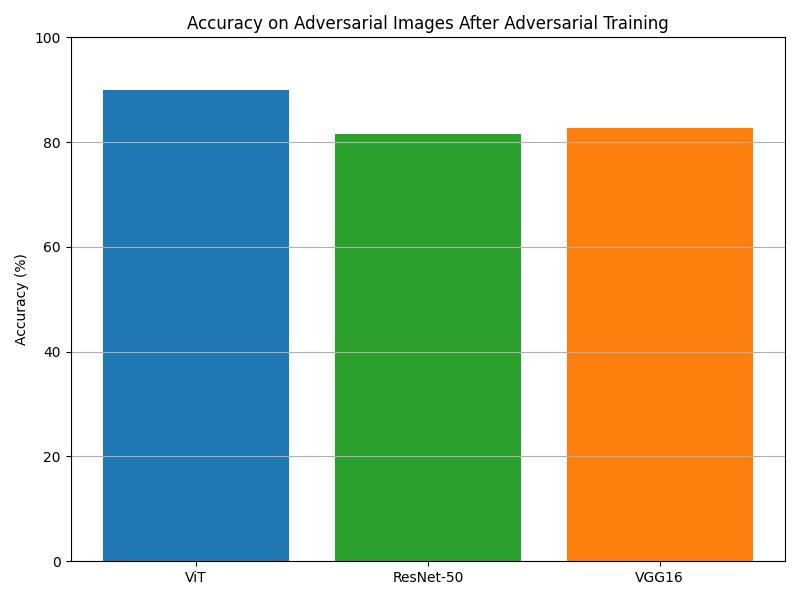}
    \caption{Post-adversarial training classification accuracy on adversarially perturbed images. }
    \label{fig:after_adv_training}
\end{figure}

To mitigate adversarial susceptibility, we performed adversarial training on all three models. This is shown in figure \ref{fig:after_adv_training}. After training with adversarial examples, we observed that ViT's accuracy on adversarial images recovered to 90.0\%, marking a substantial improvement of +62.4\%. Also, ResNet-50 improved from 70.0\% to 81.6\%, and VGG16 improved from 54.4\% to 82.6\%.

\begin{figure}
    \centering
    \includegraphics[width=0.8\linewidth]{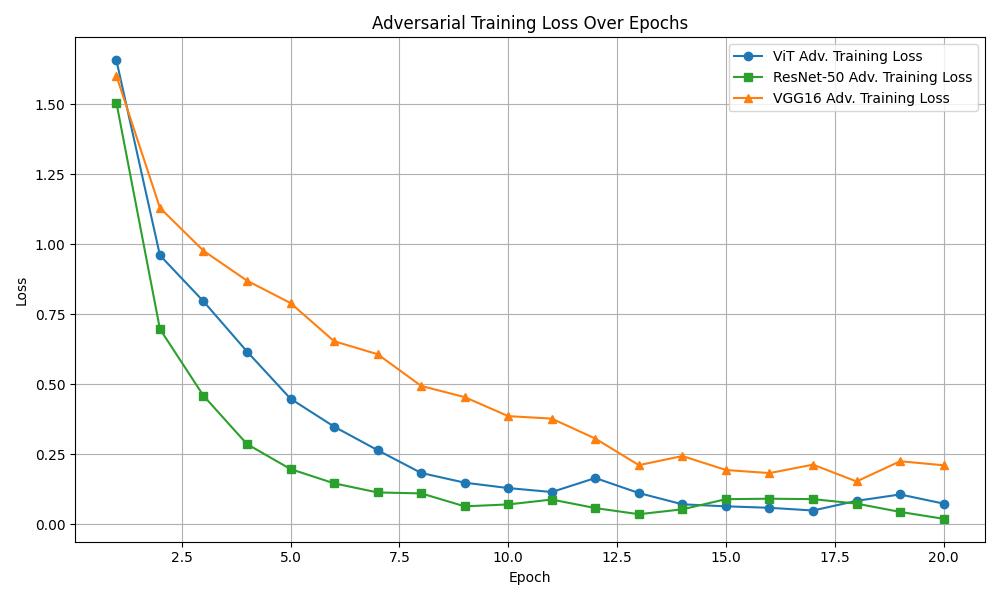}
    \caption{Adversarial training loss across 20 epochs for ViT, ResNet-50, and VGG16.}
    \label{fig:Adversarial_defence_training}
\end{figure}

Loss convergence plots during adversarial training are shown in figure \ref{fig:Adversarial_defence_training} and reveal rapid declines in training loss, especially for ViT, which converged to below 0.1 after fewer than 15 epochs. ResNet-50 and VGG16 exhibited more gradual convergence but demonstrated steady improvements. This reflects ViT’s capacity to adapt rapidly under adversarial supervision, but also indicates a need for careful regularization to avoid overfitting to adversarial patterns.

The results confirm that adversarial training is highly effective for ViT. This method not only recovers robustness but also generalizes well across previously misclassified perturbations. This finding supports the notion that initially fragile transformer models can become highly robust when trained with appropriate adversarial objectives.

In medical imaging, where a model's predictions can affect diagnoses, it's very important for models to be robust against attacks. The sharp drop in ViT’s accuracy before adversarial training shows a major risk in using unprotected transformer models in healthcare. However, after adversarial training, ViT was able to recover and perform well, proving that these models can be effective and safe for real-world medical use if properly protected.

However, our experiment used a relatively small dataset, which might not fully represent real-world scenarios in terms of diversity and complexity. Another limitation is that the computational resources were not sufficient for experiments assessing more complex architectures and larger datasets. We only included Projected Gradient Descent (PGD) for creating adversarial perturbations. PGD is a widely known and effective method for generating adversarial examples; however, we did not evaluate other adversarial attack algorithms. As a result, more comprehensive insights into model vulnerabilities could not be obtained. The final limitation is that Gaussian blur was the only preprocessing defense mechanism considered in this study. Including additional defense mechanisms would allow a more thorough evaluation of potential robustness improvements.

\section{Conclusions}

This paper presents the weaknesses of Vision Transformers (ViTs) to adversarial attacks in the context of skin disease image classification. According to this study, ViT can perform extremely well on clean images and achieves the most accurate results. However, under adversarial perturbations, ViT performs poorly and provides lower accuracy than traditional CNNs such as ResNet-50 and VGG16. The proposed adversarial training method delivers exceptionally strong performance for ViT against adversarial attacks. After adversarial training, ViT’s accuracy on adversarial images improved significantly from 27.6\% to 90.0\%. This demonstrates the model’s ability to learn robust representations when trained with properly regularized adversarial objectives.

Future research will concentrate on developing dynamic adversarial training strategies along with exploring different types of data preprocessing methods. Additionally, the effectiveness of ViT will be evaluated on larger and more diverse datasets to ensure generalizability and resilience in real-world scenarios.


\section*{Conflict of interest }

The authors declare that they have no known conflicts of interest or personal relationships that could have appeared to influence the work reported in this paper.


\bibliography{sn-bibliography}

\end{document}